\documentclass[letterpaper, 10 pt, conference]{ieeeconf}
\IEEEoverridecommandlockouts
\usepackage{cite}
\usepackage{cancel}
\usepackage{amsmath,amssymb,amsfonts,bm}
\usepackage{amsthm}
\makeatother
\usepackage{algorithmic}
\usepackage[ruled,linesnumbered,noend]{algorithm2e}
\usepackage{graphicx}
\usepackage{textcomp}
\usepackage{xcolor}
\colorlet{orange}{green!10!orange!90!}
\def\BibTeX{{\rm B\kern-.05em{\sc i\kern-.025em b}\kern-.08em
    T\kern-.1667em\lower.7ex\hbox{E}\kern-.125emX}}

\usepackage[colorlinks=true, allcolors=blue]{hyperref}
\usepackage{subcaption}
\usepackage[normalem]{ulem}
\usepackage{color}
\usepackage{xcolor}

\DeclareMathOperator*{\argmax}{arg\,max}

\newcommand{\vb}[1]{\textbf{#1}}
\newcommand{\norm}[1]{\left\lVert#1\right\rVert}

\newcommand{\mt}[1]{\mathrm{#1}}

\newtheorem{theorem}{Theorem}[section]

\newtheorem{lemma}[theorem]{Lemma}

\theoremstyle{definition}

\newtheorem{assum}{Assumption}

\begin{document}

\title{Decentralized Control of Minimalistic Robotic Swarms For Guaranteed Target Encapsulation}

\author{Himani Sinhmar and Hadas Kress-Gazit
\thanks{The authors are with the Sibley School of Mechanical and Aerospace Engineering, Cornell University, Ithaca, NY, 14853 USA. {\tt\small \{hs962,hadaskg\}@cornell.edu}. This work is supported by NSF EFMA-1935252.
}}

\maketitle

\begin{abstract}
We propose a decentralized control algorithm for a minimalistic robotic swarm with limited capabilities such that the desired global behavior emerges. We consider the problem of searching for and encapsulating various targets present in the environment while avoiding collisions with both static and dynamic obstacles. The novelty of this work is the guaranteed generation of desired complex swarm behavior with constrained individual robots which have no memory, no localization, and no knowledge of the exact relative locations of their neighbors. 
Moreover, we analyze how the emergent behavior changes with different parameters of the task, noise in the sensor reading, and asynchronous execution.
\end{abstract}

\section{Introduction}
A swarm of robots is typically composed of simple individual robots with limited capabilities.  Minimalistic swarm robotics \cite{litSurvey} emphasizes the use of simple reactive robots which use pre-programmed behaviors, similar to reflexes, without maintaining any internal state. 
The simplicity of individual robots means they can be mass manufactured and can also be scaled to micro or nano-scale. This can be particularly relevant to nanomedical applications \cite{nanobotsSurvey,Freitas2006PharmacytesAI} in which a single complex robot cannot be deployed due to space and energy constraints. 

Developing decentralized control laws for robots in a swarm that \textit{guarantee} the overall swarm behavior is a challenging task due to constraints such as limited computational power, imprecise locomotion, and the use of simple sensors. 
In this paper we present a discrete-time decentralized control algorithm for a robotic swarm with limited robot capabilities to search for and encapsulate targets in the environment, while avoiding collisions with static and dynamic obstacles. We are inspired by nanomedicine applications, such as a swarm of nano-robots searching for and encapsulating tumors by following a chemical gradient
\cite{nanoBotsNATURE,Cavalcanti_2007}. 
In addition to the control, we provide bounds on parameters that will \textit{guarantee} the swarm will achieve the task.

\noindent \textbf{Related work:} 
Work on minimalistic swarm robotics typically uses experiments or simulations to show the desired emergent behavior of the swarm given pre-programmed local rules. Using experiments with physical robots (e.g.~\cite{beckers,kube}), researchers have determined the optimal number of robots required for a swarm to complete a task; however, that work does not provide guarantees for achieving the desired behavior. 
In  \cite{moselinger} and \cite{soysal}, simulations are used as a proof of concept to evaluate the aggregation and flocking capability of a minimalistic algorithm. In \cite{martinoli}, the authors propose a probabilistic model for studying the collaborative dynamics of robots pulling sticks, which depends on the geometry but is limited to events which are constant in space and time. In \cite{onge_grid_seg,mitrano,thesis_anil,gauci_1,gauci_2,Gauci_3}, the authors either use evolutionary algorithms or an exhaustive parameter grid search over the entire space of possible controllers and provide convergence guarantees, assuming no obstacles in the environment. Moreover, in these studies, the robots are either equipped with obstacle detecting range sensors or have an infinite sensing range. 
In~\cite{Brown2016DiscoveryAE} authors use genetic algorithm with novelty search to explore the space of controllers to determine the emergent behaviors possible given a limited set of robot capabilities. 

Research in formal verification of swarms has used model checking techniques to prove properties of known swarm algorithms. 
In \cite{winfield_1,winfield_2,brambilla}, the authors use temporal logic to formally specify the emergent behaviors of a robotic swarm and verify different properties of the swarm. While it provides formal guarantees, using model checking becomes intractable as the size of the system increases. 

Recent studies have moved from minimalism and explored robots that use direct communication, broadcast information, and can learn and represent the environment. In \cite{morphogenGradients,simpleswarmforgaing} an algorithm is proposed which is based on selective broadcasting of repulsion and attraction signals among swarm agents requiring limited direct communication between robots for a given swarm task. These studies also make use of extensive simulations and experiments for emergent behavioral analysis of the swarm without providing  formal guarantees. The work in \cite{verficationEB} introduced an automatic methodology to determine whether a swarm of robots with direct communication capabilities would display an emergent behavior irrespective of the number of agents present. In \cite{mario,gazi} authors provide convergence guarantees on the emergent behavior of the swarm assuming that the robots have knowledge of the relative location of their neighbors.


\noindent \textbf{Contribution:} The novelty of this work is three fold: (i) we propose a correct-by-construction local control law for the robots that guarantees the emergence of the desired global behavior under bounds we compute on the maximum number of robots that are required for encapsulating a target, the maximum step size of a robot, and the minimum number of sensors on a robot, (ii) we consider minimalist robots that have no memory, no self-localization ability, do not know the relative location of their neighbors, no explicit communication ability, and are only equipped with omnidirectional sensors and signal emitters, and (iii) we analyze the robustness of the control algorithm to noise in the sensors and to asynchronous execution.

\section{Preliminaries}
\subsection{Environment and Robot model}

\begin{figure}[t!]
    \centering
    \includegraphics[width=0.6\linewidth]{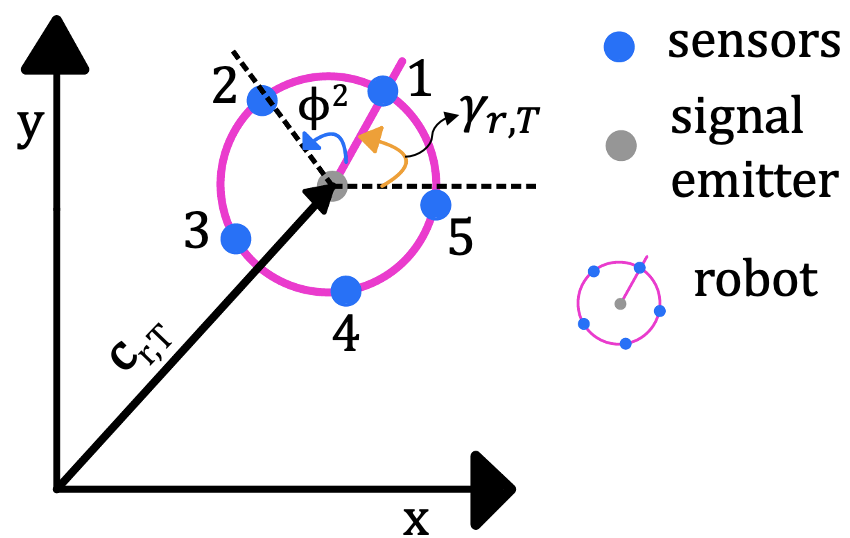}
    \caption{Robot model with $p=5$.
    }
    \label{fig:cs}
\end{figure}

\noindent \textbf{Workspace: } The robots operate in a continuous environment, $E \subseteq \mathbb{R}^2$ with a convex boundary. The environment has a fixed global frame, $\mathcal{F}$. 

\noindent\textbf{Target: }A target $g =(\mathbf{c}_{g},r_{g})$ is a disk of radius $r_{g}$ centered at $\mathbf{c}_{g}\in E$. $\mathcal{G}$ is a set of all targets contained in $E$.

\noindent \textbf{Robot: }A robot, $R = (\mathbf{c}_r,\gamma_r,r_{r},p,Z)$, is modeled as a disk of radius $r_{r}$ centered at $\mathbf{c}_r \in E$ with orientation $\gamma_r \in \mathbb{S}$, as shown in Fig. \ref{fig:cs}. 
Each robot is reactive and memoryless. It cannot localize itself and has no knowledge of the relative locations of other robots or targets. There is no explicit communication between the robots. 

The robot is controlled through rotational and translational velocities \cite{chemoController} 
in a \textit{turn-then-move} scheme
. The kinematics of a robot are described by Eq.~\eqref{robotModel} 
which is a typical unicycle model~\cite{nonholonomicMotion}. At each time step, $\theta \in \mathbb{S}$ and $d \in \mathbb{R}^+$ are the control inputs corresponding to the angle turned and the distance moved by a robot. The maximum distance a robot can move at each time step is $d_\mt{max}$. 
\vspace{-0.6em}
\begin{align} \label{robotModel} 
    \gamma_{r,T} &= \gamma_{r,T-1} + \theta \nonumber \\
    \vb{c}_{r,T} &= \vb{c}_{r,T-1} + d[\mt{cos}\gamma_{r,T} \quad \mt{sin}\gamma_{r,T}]^T
\end{align}
Each robot has $p$ omnidirectional sensors arranged on the boundary of the robot disk. $Z$ denotes the set of measurements from all sensors. 
The angle between the $k^{th}$ sensor and the robot's $x$ axis (heading direction) is denoted by $\phi^k \textrm{ }\forall k\in \{1 \cdots p\}$.

\noindent \textbf{Signal sources: }We consider three types of signals that a robot's sensor can detect: $s_g$ from a point source at the center of a target, $s_r$ from a point source at the center of the robot and $s_e$ from a line source present on the entire environment boundary. The strength of any signal $s \in \{s_g,s_r,s_e\}$ as sensed by the robot's $k^{th}$ sensor located at a distance $d^{k}_j$ from signal source $j$ is given by the function $B_s(d^{k}_j)$.  Each  signal source has a maximum influence distance $\beta_s$, such that   $B_s(d^{k}_j) = 0 \textrm{ }\mt{ } \forall d^{k}_j \geq \beta_s$. Every sensor can only sense the sum total of signal strength $z_s^k$ which it receives from all signal sources of type $s$, as defined in Eq. (\ref{totalIntesnity}). For the line source signal $s_e$, this summation becomes an integral over the boundary segment which lies inside the influence distance $\beta_e$.
\begin{equation} 
\label{totalIntesnity}    
    z_s^k = \sum_j B_s(d^{k}_j)
    \vspace{-0.6em}
\end{equation}
To model realistic sensors, we add noise to $z_s^k$; specifically, we use the noise model in~\cite{repatt}.
For example, a noise level of $15 \%$ of aggregated signal intensity is modeled as a normal distribution with a mean of 0 and a standard deviation of 0.15 as shown in Eq. (\ref{noiseAdd})
\begin{equation}
    \label{noiseAdd}
    z_s^k = (1 - n_s^k)\sum_j B_s(d^{k}_j), \, n_s^k\sim \mathcal{N}(0,0.15^2)
    \vspace{-0.6em}
\end{equation}
To ensure that $z_s^k \geq 0$, we consider a truncated normal distribution, $-1\leq n_s^k\leq1$.
For a robot, the $k^{th}$ sensor's reading consists of the tuple $(z_g^k,z_r^k,z_e^k)$.
Let $Z_g = \{z_g^1 \cdots z_g^p\}$, $Z_r = \{z_r^1 \cdots z_r^p\}$ and $Z_e = \{z_e^1 \cdots z_e^p\}$. Then the measurement set $Z = Z_g \cup Z_r \cup Z_e$.  We define $r^\mt{safe}_s \textrm{ }\forall s \in \{g,r,e\}$ as the distance each robot must maintain from a source at all times. 

\begin{figure}[t!]
    \centering
    \includegraphics[width=0.8\linewidth]{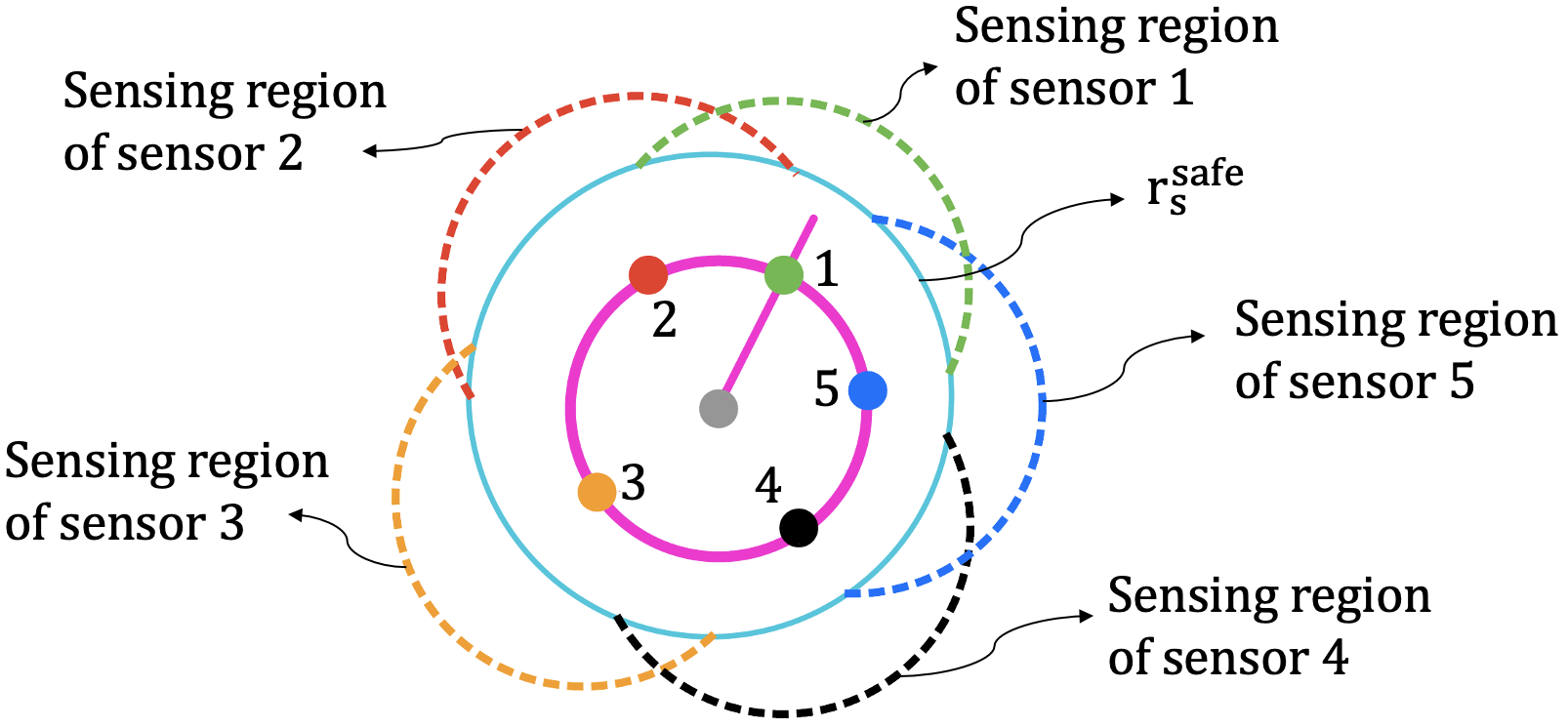}
    \caption{Possible asymmetric placement of sensors for $p=5$. The sensing region of a sensor is equivalent to the influence region of a source. As can be seen, any source $s$, located at a distance of $r_s^{\mt{safe}}$ from the robot's center (cyan colored circle) will be inside the sensing region of at least one sensor.  
    }
    \label{fig:sensorArrange}
\end{figure}

We assume the sensors are arranged on the robot's boundary such that at least one sensor is in the influence region of a source, $s$, when the robot center is $r_s^{\mt{safe}}$ away from $s$. Note that we do not assume symmetric placement;  Fig. \ref{fig:sensorArrange} depicts one such valid, asymmetric placement.
For the ease of exposition, in this paper we use a separation angle of $2\pi /p$ between sensors and indicate what change needs to be made in the case of asymmetric placement.

\noindent \textbf{Target encapsulation:}
Let $\mathcal{A}_g = (\mathbf{c}_g,r^\mt{safe}_g,r^\mt{encap}_g)$ be an annular region between two concentric circles of radius $r^\mt{safe}_g$ and $r^\mt{encap}_g$ centered at $\mathbf{c}_g$ such that $r^\mt{encap}_g > r^\mt{safe}_g$. The target $g \in \mathcal{G}$ is said to be \textbf{encapsulated} if the total number of robots currently present in the annular region, $\mathcal{A}_g$, is $n_g$. A robot is considered to be in the annular region of a target $g$ if, 
\begin{equation} \label{trappingCondition}
    r^\mt{safe}_g<\norm{\vb{c}_r-\vb{c}_g}\leq r^\mt{encap}_g
\end{equation}

\begin{assum}
\label{only1TarSensed}
Any two targets are at least ($2\beta_g+2r_r+\epsilon$) units apart, where $\epsilon$ is a small positive number. This ensures that a robot can sense at most one target at a time.
\end{assum}

\begin{assum}
\label{AstopFlag}
When a target is encapsulated, it stops emitting a signal and emits a single burst of a shut off signal. The influence distance of this signal is limited to $r^\mt{encap}_g$ , and we assume that robots within the influence region, i.e. in the annular region $\mathcal{A}_g$, set $\bm{u}=0$ thereafter but keep emitting their signal to ensure no collisions with other moving robots. This assumption emulates the behavior of nanorobots which would wrap around the tumor's surface to destroy it. 
\end{assum}

\begin{assum}
\label{AdistKnown}
The inverse of the signal function $B_s$ exists and is known to the robots, i.e. given a signal strength reading of source $s$ for the $k^{\mt{th}}$ sensor, a robot can compute the distance $d^{k}_s = B_s^{-1}(z^k_s)$. Furthermore, we assume that the signal strength strictly decreases with the radial distance from a source. Since each sensor only senses an aggregated signal, the computed distance $d^{k}_s$ is typically not the true distance to a single source. 
\end{assum}

\section{Problem Formulation}
Consider $m$ targets in an environment $E$. Given the user-provided safe distance $r^{\mt{safe}}_s$ a robot $R$ needs to maintain from a source, the number of robots $n_g$ needed to encapsulate each target $g \in \mathcal{G}$, the total number of robots  $n$ such that $n \geq \sum_{g \in \mathcal{G}} n_g$, and the total number of sensors $p$ on a robot, find a control law $\bm{u}(t)$ that will result in all targets $\mathcal{G}$ being encapsulated while ensuring the robots maintain $r^{\mt{safe}}_s$.

In the following we derive such a control law while providing bounds on the parameters $p$, $d_\mt{max}$, $\beta_r$, $r^\mt{encap}_g$ and $n_g$ which guarantee collision free encapsulation behavior. 

\section{Approach}
\label{approach}
We design the control of a robot as a combination of three behaviors: (i) random walk when no target is sensed, (ii) moving towards a target when sensing one, and (iii) collision avoidance with the other robots, the targets and the environment boundary. The robot transitions between behaviors based on the signal thresholds $I^\mt{safe}_s\textrm{ } \forall s \in \{s_g,s_r,s_e\}$.
In Section $\ref{intensity_rsafe}$ we define virtual sources and use them later to compute $I^\mt{safe}_s$ based on $r^\mt{safe}_s$.
In Sections \ref{tarAtt} and \ref{collAvo} we find control parameters for a single robot that guarantee it will move towards a target while avoiding collisions.

\subsection{Virtual Sources}
\label{intensity_rsafe}
\begin{figure}[b!]
    \centering
    \includegraphics[width=0.8\columnwidth]{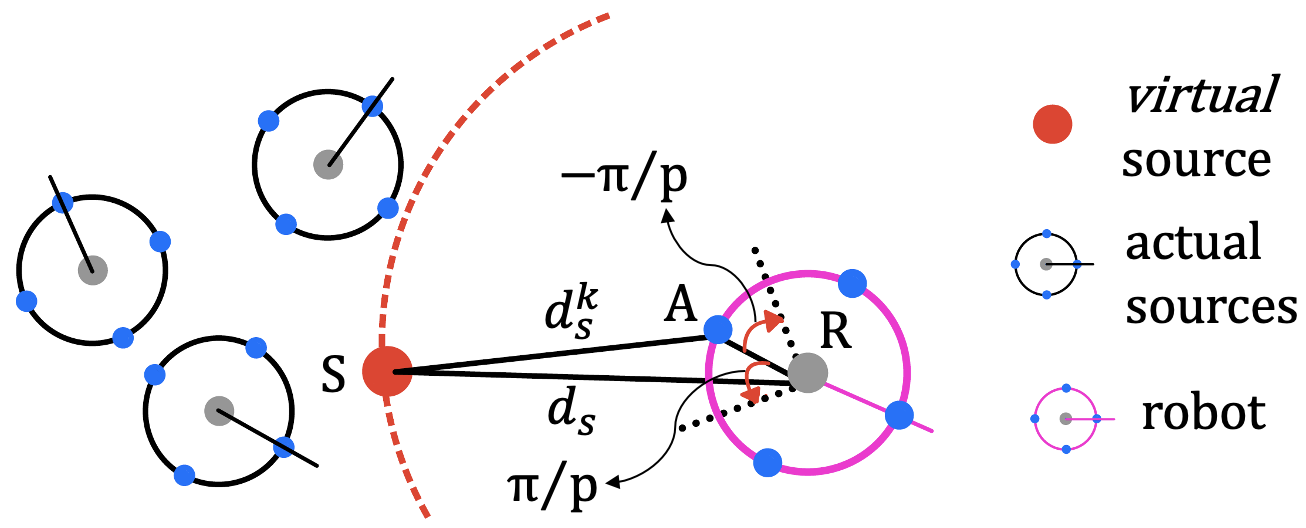}
    \caption{\textit{Virtual} source $S$ located  $d_s^k$ from the  $k^{th}$ sensor and $d_s$ from robot center $R$ such that $\angle{SRA} \in  [ -\pi/p,\textrm{ }\pi/p]$.
    }
    \label{fig:effectSource}
\end{figure}

Since a robot's sensor measures only the total signal strength from nearby sources, it can neither figure out the relative location of the sources nor the total number of sources it is currently sensing.  The same signal strength could correspond to a single source nearby or a cluster of sources farther away. Therefore, we define a \textit{virtual} source at a radial distance from the sensor such that the signal strength from this source is equivalent to the total signal strength sensed by the sensor.  
Fig. \ref{fig:effectSource} shows a single robot centered at point $R$ and a \textit{virtual} source centered at point $S$.
The distance $d_s^k = B_s^{-1}(z^k_s)$ is known from the sensor reading $z^k_s$ of the $k^{th}$ sensor (point $A$ in Fig. \ref{fig:effectSource}), and $S$ is located on a circle centered at point $A$ with radius $d_s^k$. From the geometry of $\triangle{SRA}$ we have,
\begin{equation}
    \label{virtualSource}
    \overline{RS} = r_r\mt{cos}(\angle{SRA}) + \sqrt{(d_s^{k})^2-r_r^2\mt{sin}^2(\angle{SRA})} 
\end{equation}

Assuming sensor $k$ is receiving the strongest signal from source $s$, i.e. $z^k_s \geq z^l_s, \forall l\neq k$, we are interested in finding the shortest possible distance $d_s$ between the robot and the \textit{virtual} source. We will use that to determine the maximum distance the robot can safely move.
Now, if the \textit{virtual} source is placed such that $\angle{SRA} > \pi/p$ in either direction of $RA$, then the maximum signal reading would be seen at $(k\pm1)^{\mt{th}}$ sensor and not at this sensor. This restricts the range of possible directions in which the \textit{virtual} source can be located with respect to the robot's center to $\angle{SRA} \in  [ -\pi/p,\textrm{ }\pi/p]$. 
It can then be seen that the source will be closest to the robot's center when $\angle{SRA} = \pm\pi/p$ and farthest when $\angle{SRA}=0$. Substituting $\angle{SRA}=\pi/p$ and $\overline{RS}=d_s$ in the above equation we have,
\vspace{-0.5em}
\begin{equation} \label{dsk}
   d_s = r_r\mt{cos}(\pi/p) + \sqrt{(d_s^{k})^2-r_r^2\mt{sin}^2(\pi/p)}
\end{equation}
\noindent \textbf{Asymmetric sensor placement:} Replace $\pi/p$ in Eq.~\eqref{dsk} with half of the maximum angle that the $k^{\mt{th}}$ sensor makes with either of its adjacent sensors.

\subsection{Target attraction }
\label{tarAtt}
\begin{figure}[h!]
    \centering
    \includegraphics[width=1\columnwidth]{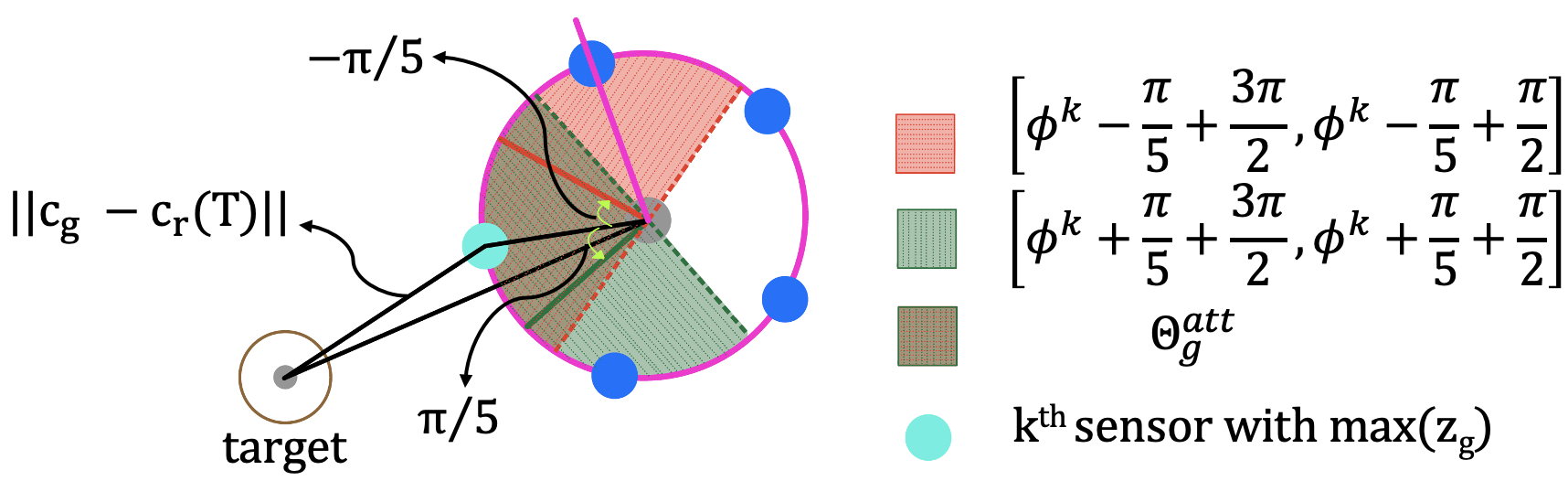}
    \caption{The direction of motion $\theta$ for a robot should be within the angular range $\Theta^\mt{att}_g$ for it to move towards a target.
    }
    \label{fig:tarAtt}
\end{figure}

We use the Lyapunov stability theorem \cite{Khalil} to find the control parameters $d$ and $\theta$ such that a robot moves towards a target $g \in \mathcal{G}$ when sensing it. 
Let $V = \norm{\mathbf{c}_g-\vb{c}_{r,T}}^2 > 0$ be the candidate Lyapunov function, then 
$\mathbf{c}_g$ is stable if $V_{T+1}\leq V_T$, that is, $\norm{\mathbf{c}_g-\vb{c}_{r,T+1}}^2 \leq \norm{\mathbf{c}_g-\vb{c}_{r,T}}^2 $.
Using Eq. (\ref{robotModel}),
\begin{align*}
    \norm{\mathbf{c}_g-\vb{c}_{r,T}-\mathbf{u}}^2 &\leq \norm{\mathbf{c}_g-\vb{c}_{r,T}}^2
\end{align*}
Let $\gamma_g$ be the angle between the vectors $(\mathbf{c}_g-\vb{c}_{r,T})$ and $\mathbf{u}$, and $d = \norm{\mathbf{u}}$ then,
\begin{equation}
    \cancel{\norm{\mathbf{c}_g-\vb{c}_{r,T}}^2}+d^2-2d\norm{\mathbf{c}_g-\vb{c}_{r,T}}\mt{cos}\gamma_g \leq \cancel{\norm{\mathbf{c}_g-\vb{c}_{r,T}}^2}
\end{equation}
\begin{equation}
\label{goalAttCond}
    d^2-2d\norm{\mathbf{c}_g-\vb{c}_{r,T}}\mt{cos}\gamma_g  \leq 0
\end{equation}
Ignoring the unlikely, perfectly symmetric scenario where 2 sensors receive the maximum intensity from a target, let $k$ be the index of the sensor such that $z^k_g > z^l_g, \forall l\neq k$. Then 
the direction of the target with respect to the robot's center is within the angular range $[\phi^k-\pi/p,\textrm{ }\phi^k+\pi/p]$ as explained in Section \ref{intensity_rsafe}. This implies that $\gamma_g \in [\theta - (\phi^k-\pi/p),\textrm{ }\theta - (\phi^k+\pi/p)]$ where $\theta$ is the direction of the robot motion.
The necessary condition to satisfy Eq. (\ref{goalAttCond}) is that $\mt{cos}\gamma_g \geq 0 $, that is, $\gamma_g \in [3\pi/2,\textrm{ }2\pi]\cup [0,\textrm{ }\pi/2]$. Using these two conditions, the angular range of possible control directions $\theta$ is given by Eq. (\ref{goalAttAng}). Fig. \ref{fig:tarAtt} shows an example of the angular range $\Theta_g^{\mt{att}}$ for a robot with $p = 5$ as computed by its sensor closest to the target.
\begin{equation} \label{goalAttAng}
    \Theta_g^\mt{att} = [\phi^k + \pi/p + 3\pi/2, \textrm{ } \phi^k - \pi/p + \pi/2]
\end{equation}
Let $d_g$ be the estimate of $\norm{\mathbf{c}_g-\vb{c}_{r,T}}$ obtained using $z^k_g$, then from Eq. (\ref{goalAttCond}), $d \leq 2d_g\mt{cos}\gamma_g$.
In Algorithm \ref{algo:ca1}, \textbf{\texttt{DistAttractTarget}} computes the maximum possible value of $d$ such that, $0 \leq d \leq \mt{min}\{2d_g\mt{cos}\gamma_g,\textrm{ }d_{\mt{max}}\}$.
\noindent \textbf{Asymmetric sensor placement:} Replace $-\pi/p$ and $\pi/p$ in Eq.~\eqref{goalAttAng} with half of the angle that the $k^{\mt{th}}$ sensor makes with sensor $k-1$ and sensor $k+1$ respectively (assuming counterclockwise ordering of sensors as shown in Fig. \ref{fig:cs}).

\subsection{Collision Avoidance}
\label{collAvo}
To avoid collisions with a source $s \in \{g,r,e\}$, we set signal thresholds $I_s^\mt{safe}$ such that the robot triggers collision avoidance behavior before the distance between a source and a robot's center, as estimated from a sensor's reading, is equal to $r^\mt{safe}_s$.
Since $r^\mt{safe}_s$ is defined between the robot's center and the source we set $d_s = r^\mt{safe}_s$ in Eq. (\ref{dsk}), to obtain $d_s^k = \sqrt{(r_s^\mt{safe})^2 + r_r^2 - 2r_rr_s^\mt{safe}\mt{cos}(\pi/p)}$. Furthermore, to account for a scenario where the distance between the source and the robot is just marginally greater than $r^{\mt{safe}}_s$, we add the maximum distance $d_\mt{max}$ the robot can move at this time step to $d_s^{k}$. Then
the threshold strength $I_s^\mt{safe}, s\in\{g,r,e\}$ is given by Eq. (\ref{I_safe}).
\vspace{-0.4em}
\begin{equation} \label{I_safe}
I^\mt{safe}_s = B_s\bigg(d_\mt{max} + \sqrt{(r_s^\mt{safe})^2 + r_r^2 - 2r_rr_s^\mt{safe}\mt{cos}(\pi/p)} \textrm{ }\bigg) 
\end{equation} 
\noindent \textbf{Asymmetric sensor placement:} Replace $\pi/p$ in Eq. ~\eqref{I_safe} with half of the maximum angle between two adjacent sensors on the robot. 

\noindent\textbf{Collision avoidance with a static obstacle: }
Let $\mathbf{c}_o^\mt{static}$ be the location of a static obstacle (target or boundary). To avoid collision, the robot's motion at time step $T$ should be such that it does not move towards the obstacle, that is $\norm{\mathbf{c}_o^\mt{static}-\vb{c}_{r,T+1}} \geq \norm{\mathbf{c}_o^\mt{static}-\vb{c}_{r,T}}$.

Let $k$ be the index of the sensor receiving the maximum intensity from the static source $s \in \{g,e\}$ such that $z^k_s \geq I_s^{safe}$ and  $z^k_s>z^l_s$, $\forall l\neq k$. Then the direction of the obstacle with respect to the robot's center is within the angular range $[\phi^k-\pi/p,\textrm{ }\phi^k+\pi/p]$ as explained in Section \ref{intensity_rsafe}. Let $\gamma_o$ be the angle between the vectors $(\mathbf{c}_o^{\mt{static}}-\vb{c}_{r,T})$ and $\mathbf{u}$, that is $\gamma_o \in [\theta - (\phi^k-\pi/p),\textrm{ }\theta - (\phi^k+\pi/p)]$. Now, using Eq. (\ref{robotModel}) to simplify the collision avoidance constraint we get,
\begin{align} \label{colAvoAng}
    d^2-2d\norm{\mathbf{c}_o^{\mt{static}}-\vb{c}_{r,T}}\mt{cos}\gamma_o \geq 0
\end{align}
If $\gamma_o$ is chosen such that $\mt{cos} \gamma_o \leq 0$, then Eq. (\ref{colAvoAng}) is always satisfied. Using both constraints, the angular range for $\theta$ to avoid static obstacles is given by Eq. (\ref{angrangeAvo}). 
\begin{equation} \label{angrangeAvo}
    \Theta_s^\mt{avo} = [\phi^k + \pi/p + \pi/2, \textrm{ } \phi^k - \pi/p + 3\pi/2] \\ 
\end{equation}
\noindent \textbf{Asymmetric sensor placement:} Replace $-\pi/p$ and $\pi/p$ in Eq.~\eqref{angrangeAvo} with half of the angle that $k^{\mt{th}}$ sensor makes with sensor $k-1$ and sensor $k+1$ respectively (assuming counterclockwise ordering of sensors as shown in Fig. \ref{fig:cs}).


\noindent\textbf{Collision avoidance with other robots: }
In the case of dynamic obstacles (i.e. other robots), the condition in Eq. (\ref{colAvoAng}) would not be sufficient for collision avoidance, since the other robot can move toward the robot. Furthermore, a robot might be surrounded by multiple moving robots, so the control parameters should be chosen such that it avoids all of the nearby robots. 

\begin{figure}[h!]
    \centering
    \includegraphics[width=0.7\linewidth]{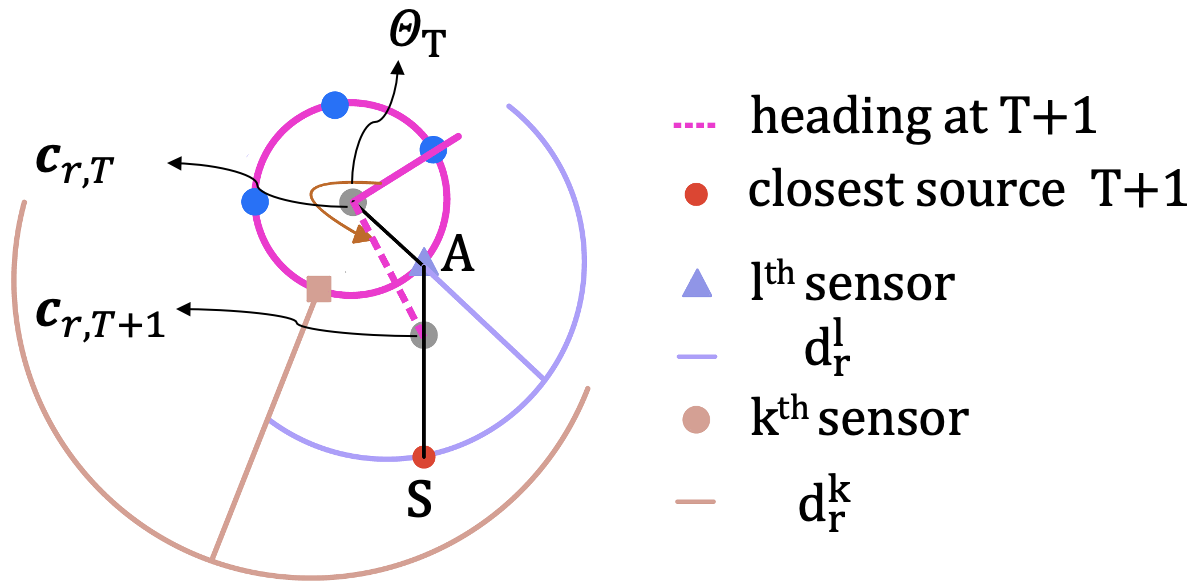}
    \caption{The distance to move in $\theta$ direction is computed using the reading from $k^{th}$ and ${l^{th}}$ sensor.
    \label{fig:dynColl}}
    \vspace{-1em}
\end{figure}

For any sensor $k$, its virtual source is located at a radial distance of $d_r^k = B_r^{-1}(z_r^k)$ within the angular range $[\phi^k-2\pi/p, \textrm{ } \phi^k+2\pi/p]$ as shown in Fig. \ref{fig:dynColl}. 
At time step $T$, given the direction of motion $\theta$ (shown by the dashed magenta line), let $k$ and $l$ be the indices of the sensors that are closest to $\theta$ i.e. $\phi^k < \theta < \phi^l$ such that $d_r^l < d_r^k$.
Then the distance that the robot can move is chosen such that it maintains a safe distance of $r_r^\mt{safe}$ from neighboring moving robots after moving $d$ units in the direction of motion. At $T+1$, the closest virtual source to the robot is at $\vb{S}$. The maximum distance that this virtual source could have moved at $T$ is $d_\mt{max}$.
To ensure safety, $\norm{\vb{c}_{r,T+1}-\vb{S}} \geq r_r^{\mt{safe}} + d_{\mt{max}}$. Using the geometry of $\triangle{\vb{c}_{r,T}A\vb{c}_{r,T+1}}$, we have
\begin{align}
    d_r^l - \sqrt{d^2+r_r^2-2dr_r\mt{cos}(\phi^l-\theta)} &\geq r_r^{\mt{safe}} + d_{\mt{max}}
\end{align}
\vspace{-2.2em}
\begin{multline}
    \label{avoObsGivenTh}
    0 \leq d \leq r_r \mt{cos}(\phi^l-\theta) \\
    + \sqrt{(d_r^l - r_r^{\mt{safe}} - d_{\mt{max}})^2 - r_r^2 \mt{sin}^2(\phi^l-\theta)}
\end{multline}
In Algorithm \ref{algo:ca1}, the function \textbf{\texttt{DistAvoDynObs}} computes the maximum possible value of $d$ from Eq. (\ref{avoObsGivenTh}).
\vspace{-0.3em}
\subsection{Control for Each Robot}
\label{controlRob}
\textbf{Algorithm} (\ref{algo:ca1}) describes the control generation for a robot in the swarm. If the total signal strength received by a robot from static sources $s \in \{g,e\}$ is greater than the preset safe threshold, i.e. max$(Z_s)\geq I_s^\mt{safe}$, the robot finds a direction of motion ($\theta \in \Theta_s^{\mt{avo}}$) that maximizes the possible distance, $d$,
such that it moves away from the source $s$ while avoiding nearby moving robots (lines 2-3). 

When the robot senses a target $g \in \mathcal{G}$ such that the maximum signal strength received is less than the safety threshold $I_g^\mt{safe}$, it moves towards the target while avoiding collisions with nearby robots. The parameters $d$ and $\theta$ for this behavior are chosen based on Section \ref{tarAtt} (lines 4-6). 

If the computed distance $d$ is zero, the robot chooses the control parameters based on the reading of sensor $k$ receiving the minimum signal strength $z_r^k$ (lines 7-10), as that is the safest direction to move in. When the robot is outside the influence of all targets, it performs a random walk while avoiding collisions with nearby robots (lines 11-17).
\begin{algorithm}[h!]
\SetKwInOut{Input}{Input}
\SetKwInOut{Output}{Output}
\SetKwProg{Initialization}{Initialization}{}{}
\Input{$Z$, $B_s$, $p$, $r_s^{\mt{safe}}$, $\forall s \in \{r,g,e\}$}
\Output{$d$, $\theta$}
 \tcp{compute $\Theta_g^\mt{att}$, $\Theta_e^\mt{avo}$, $\Theta_g^\mt{avo}$}
\If{\texttt{max}$(Z_s) \geq I_s^\mt{safe}$ $, s \in \{g,e\}$}{
    $\theta$ = $\argmax\limits_{\varphi \in \Theta_s^{\mt{avo}}}$ \texttt{DistAvoDynObs}$(Z_r,B_r,\varphi,r_r^{\mt{safe}})$\\
    $d$ = \texttt{DistAvoDynObs}$(Z_r,B_r,\theta,r_r^{\mt{safe}})$\\
}
\ElseIf{
    $0<\mt{max} (Z_g)<I^{\mt{safe}}_g$}{
    $\theta$ = $\argmax\limits_{\varphi \in \Theta_g^{\mt{att}}}$ \texttt{DistAvoDynObs}$(Z_r,B_r,\varphi,r_r^{\mt{safe}})$\\
    $d$ = \texttt{min}\big(\texttt{DistAvoDynObs}$(Z_r,B_r,\theta,r_r^{\mt{safe}})$, \texttt{DistAttractTarget}$(Z_g,B_g,\theta,r_g^{\mt{safe}})$\big)\\
    \If{$d=0$}{
        $k= \texttt{argmin}(Z_r)$ \\
        $\theta$ = $\phi^k$\\
        $d$ = \texttt{DistAvoidRobots}$(Z_r,B_r,\theta,r_r^{\mt{safe}})$\\
    } 
}
\Else{
    $\theta$ = \texttt{randsample}($[0,\textrm{ } 2\pi]$)\\
    $d$ = \texttt{DistAvoDynObs}$(Z_r,B_r,\theta,r_r^{\mt{safe}})$\\
    \If{$d=0$}{
        $k= \texttt{argmin}(Z_r)$ \\
        $\theta$ = $\phi^k$\\
        $d$ = \texttt{DistAvoidRobots}$(Z_r,B_r,\theta,r_r^{\mt{safe}})$\\
    } 
}
\caption{Control algorithm for a robot}
\label{algo:ca1}
\end{algorithm}
\vspace{-0.4em}
\section{Safety and Liveness Guarantees}
\label{proof}
In this section we assume noiseless sensors and derive constraints on different parameters, that if satisfied, guarantee that there are no collisions, the robots are never stuck in a deadlock and all targets in the environment are encapsulated. In Section \ref{ba} we analyze how sensor noise and asynchronous control execution affect these guarantees and lead to interesting emergent behaviors. 

\subsection{Safety: Collision avoidance}
\begin{lemma}
\label{closestObsLemma}
The estimate of the relative distance $d_s$ between a \textit{virtual} source and robot's center in Eq. (\ref{dsk}) is always less than or equal to the distance from the actual closest source.
\end{lemma}

\begin{proof}
A sensor receives the sum total of signal strengths from all nearby sources, which is always greater than or equal to the signal strength from a single source because $B_s(d)$ is always positive. From Assumption (\ref{AdistKnown}) and Eq. (\ref{totalIntesnity}), the radial distance $d_s^k$ of the \textit{virtual} source from the $k^\mt{th}$ sensor decreases as the sensor reading $z_s^k$ increases. Hence, the radial distance $d_s^k$ is  always equal to (if there is a single source) or less than (if there are multiple nearby sources) the radial distance from the actual closest source. As detailed in Section \ref{intensity_rsafe} and Eq. (\ref{dsk}), a robot always chooses the minimal possible radial distance $d_s$ from the virtual source.
\end{proof}

\begin{lemma}
\label{collAvoThm}
If every robot in the swarm implements a local behavior, as outlined in the Algorithm \ref{algo:ca1}, such that the following requirements hold, then a robot always maintains a given safe distance $r_s^{\mt{safe}}$ from a source $s \in \{g,r,e\}$: 
\begin{enumerate}
\label{require}
    \item The influence distance of a source $\beta_s \geq
    B^{-1}_s(I_s^\mt{safe})$
    \item In the initial state, a robot is at least $r_s^{\mt{safe}}$ units away from every source
\end{enumerate}
\end{lemma}
\begin{proof}
Condition (1) ensures that the maximum influence distance of a source is at least $r_s^{\mt{safe}}$. Using Lemma \ref{closestObsLemma} and Eq. (\ref{I_safe}), collision avoidance behavior (\ref{collAvo}) is always triggered for a robot before the distance between a source and a robot's center, as estimated from a sensor's reading, is equal to $r^\mt{safe}_s$.
As elaborated in Section \ref{collAvo}, a robot always either moves in a direction with no obstacles or moves a distance $d$ such that a minimum distance of $r^{\mt{safe}}_s$ is maintained from the \textit{virtual} source.
Condition (2) enforces that the swarm is safe at the initial time step.
\end{proof}

\subsection{Absence of deadlocks}
\begin{figure}[t!]
\centering
\includegraphics[width=0.6\columnwidth]{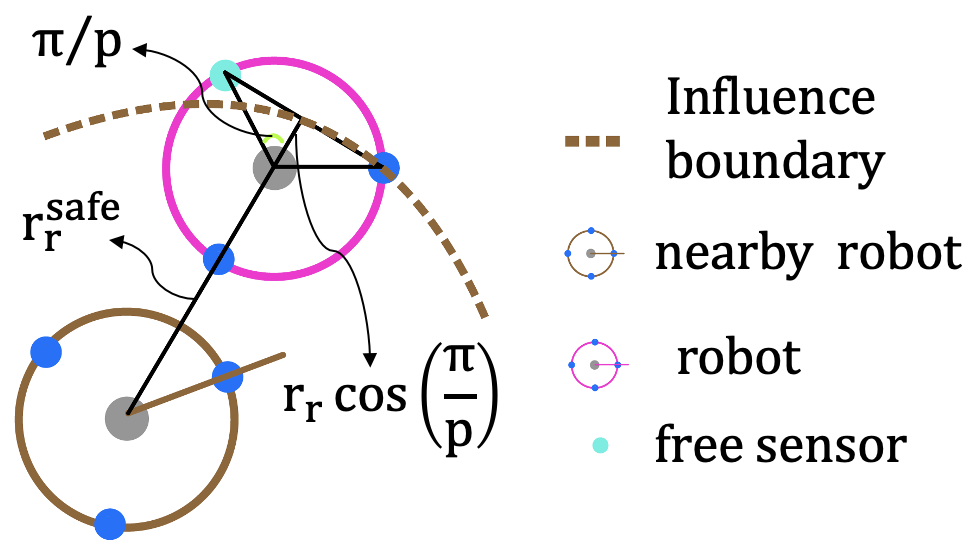}
\caption{When two robots are $r_r^{\mt{safe}}$ distance apart at least one sensor is outside the influence region $\beta_r$.}
\label{fig:senseRad}
\end{figure}

\begin{figure}[t!]
    \centering
    \includegraphics[width=0.48\linewidth]{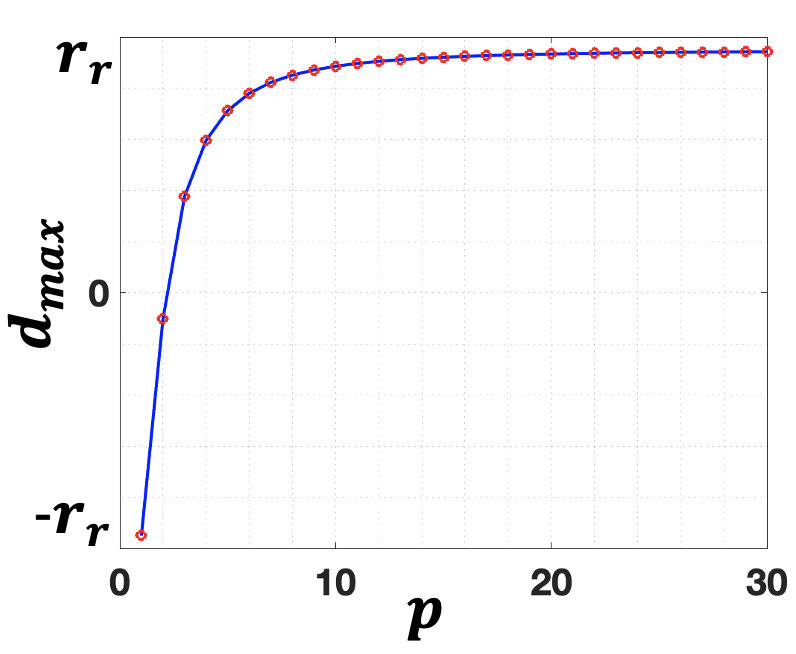}
    \caption{$d_{\mt{max}}$ as a function of $p$ for equally spaced sensors. 
    \label{fig:dmaxP}}
\end{figure}

A deadlock occurs in a swarm if two or more robots are in a configuration where none of them can move, i.e. $d=0$. 
\begin{lemma}
\label{noDeadLemma}
No deadlocks can occur in a swarm if,
\begin{enumerate}
\label{require_dead}
    \item The total number of robots in the swarm is such that they can be placed in a configuration where any two robots are at least ($\beta_r + r_r$) units apart. 
    \item The maximum influence distance of a robot's source $\beta_r$  satisfies $B_r^{-1}(I^\mt{safe}_r) + d_{\mt{max}} < \beta_r < r_r^{\mt{safe}}+r_r\mt{cos}(\pi/p)$
    \item The maximum distance a robot can move in a time step 
    $d_{\mt{max}} < \frac{r_r^{\mt{safe}}+r_r\mt{cos}(\pi/p) -\sqrt{(r_r^\mt{safe})^2 + r_r^2 - 2r_rr_r^\mt{safe}\mt{cos}(\pi/p)}}{2}$
    \item The total number of sensors on a robot is $p \geq 3$
\end{enumerate}
\end{lemma}
\begin{proof}
To ensure safety, the influence region of a robot's source should be large enough so that Lemma \ref{collAvoThm} is satisfied. In a scenario where a robot is marginally outside $\beta_r$ of a nearby robot and moves $d_{\mt{max}}$, we get $\beta_r > B_r^{-1}(I^\mt{safe}_r) + d_{\mt{max}}$. Furthermore, the influence distance should be small enough such that when two robots are at a relative distance of $r_r^{\mt{safe}}$, at least one sensor on each robot is outside the influence region of the other robot.
This, along with condition (1), ensures that there always exists at least one robot in the swarm which has $\mathbf{u}\neq 0$ in an obstacle free direction. 

Fig. \ref{fig:senseRad} shows two robots that are $r_r^{\mt{safe}}$ apart. From geometry, we get $\beta_r < r_r^{\mt{safe}} +r_r\mt{cos}(\pi/p)$. These constraints on $\beta_r$ give us an upper bound on the maximum step size of a robot,
\begin{equation}
\label{boundOnSenseRad}
    B_r^{-1}(I^\mt{safe}_r) + d_{\mt{max}}< r_r^{\mt{safe}} + r_r\mt{cos}(\pi/p)
\end{equation}
Using Eq. (\ref{I_safe}) we have,
\begin{multline}
     2d_{\mt{max}} + \sqrt{(r_r^\mt{safe})^2 + r_r^2 - 2r_rr_r^\mt{safe}\mt{cos}(\pi/p)} \\
     < r_r^{\mt{safe}} + r_r\mt{cos}(\pi/p)  
\end{multline}
\vspace{-2.5em}
\begin{multline}
    \label{dmax}
    d_{\mt{max}} < \frac{r_r^{\mt{safe}}+r_r\mt{cos}(\pi/p)}{2}\\ -\frac{\sqrt{(r_r^\mt{safe})^2 + r_r^2 - 2r_rr_r^\mt{safe}\mt{cos}(\pi/p)}}{2}
\end{multline}
Fig. \ref{fig:dmaxP} shows how $d_{\mt{max}}$ (as a function of $r_r$) changes with the number of sensors $p$. As can be seen from Eq.~\eqref{dmax}, as $p$ grows, $d_{\mt{max}}$ approaches $r_r$. Furthermore, we require $p \geq 3$ otherwise $d_{\mt{max}} < 0$ for any given $r^\mt{safe}_r > 0$. 
\end{proof}
\noindent \textbf{Asymmetric sensor placement:} Replace $\pi/p$ in Eq. ~\eqref{boundOnSenseRad} with half of the maximum angle between two adjacent sensors on a robot. 

\begin{figure}[t!]
    \centering
    \includegraphics[width=0.6\linewidth]{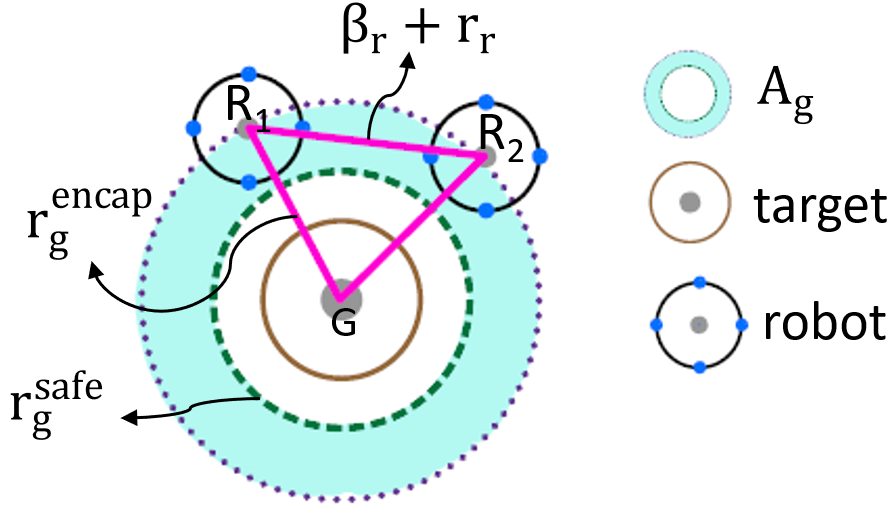}
    \caption{Geometry of a target and two robots to compute an upper bound on $n_g$. }
    \label{fig:totRobs}
\end{figure}

\subsection{Liveness: Encapsulating all targets}
\begin{lemma}
\label{rwProp}
    A robot performing a random walk in a bounded environment will always eventually explore the entire area \cite{popov_2021}.
\end{lemma}

\begin{lemma}
\label{liveLemma}
    For any random initial condition such that,
\begin{enumerate}
    \item the number of robots $n_g$ required to encapsulate a target $g$ satisfies,
    \vspace{-0.4em}
    \begin{equation}
        \label{totalRobs}
        n_g \leq \frac{2\pi}{\mt{cos}^{-1}\bigg(1 - \frac{(\beta_r + r_r)^2}{2(r^\mt{encap}_g)^2}\bigg)} = n_0
    \end{equation}
    \item the outer radius of the annular region $\mathcal{A}_g$, $r^{\mt{encap}}_g \geq r_g^{\mt{safe}} + 2d_{\mt{max}}$.
\end{enumerate}
all $g\in\mathcal{G}$ will eventually be encapsulated.
\end{lemma}
\begin{proof}
The maximum number of robots, $n_0$, that can be present simultaneously in the annular region $\mathcal{A}_g$ to satisfy Eq. (\ref{trappingCondition}) should be such that they are outside each other's influence region. 
That is, the minimum relative distance between any two consecutive robots is $\beta_r + r_r$. This, along with condition (2), ensures that there exists a configuration where the robots encapsulating a target are in the annular region without any chattering. 
From Fig. \ref{fig:totRobs}, $\angle{R_2GR_1} = cos^{-1}\bigg(1 - \frac{(\beta_r + r_r)^2}{2(r^\mt{encap}_g)^2}\bigg)$ is given by the cosine rule of triangles. We define $n_0 = \frac{2\pi}{\angle{R_2GR_1}}$. Since in the annular region a robot oscillates between getting attracted to a target and maintaining a distance of $r_g^{\mt{safe}}$ from it, the upper bound on $r^{\mt{encap}}_g$ is such that Eq. (\ref{trappingCondition}) can be satisfied  despite these oscillations. 

Let the number of robots specified to encapsulate a target $n_g $ be less than $n_0$. At any time step $T$, if there are less than $n_g$ robots in $\mathcal{A}_g$, a robot in the influence of a target always finds either an obstacle free direction to move towards the target or it finds a non-zero distance to move in the direction of the sensor receiving the minimum signal from nearby robots (Lemma \ref{noDeadLemma}). 
This behavior leads to an increment in the Lyapunov function of the robot but ensures that the robot does not get stuck in a local minima caused by an obstacle 
between itself and the target.
When $n_g > n_0$, there is a dynamic equilibrium of robots near the target such that at least $n_0$ robots are almost always present in $\mathcal{A}_g$. 

From assumption (\ref{AstopFlag}), when a target is encapsulated, it stops emitting a signal and all the robots in $\mathcal{A}_g$ stop moving. The robots that were outside $\mathcal{A}_g$ but inside the target's influence region will transition into a random walk behavior. It follows from Lemma \ref{rwProp} and Lemma \ref{noDeadLemma} that a swarm will always eventually encapsulate all the targets.
\end{proof}

\section{Analysis of Swarm Behavior}
\label{ba}
In this section, we investigate how the parameters $p$, $n_g$, and noisy sensors affect the emergent behavior of the swarm. 
We used three metrics to compare the emergent behavior: the total time for encapsulating all targets $g\in\mathcal{G}$, the cumulative path length traveled by  all robots, and the probability of target encapsulation while ensuring no collisions. 

In the following, the environment consists of one target and the number of robots is $n = 10$. We fixed the total simulation time to be 3000 time steps and ran 100 simulations for each data point with the same initial conditions. This was done to study the effect of the parameters on the emergent behavior by keeping other conditions constant. The randomness in each simulation is due to the random choice of $\theta$. Moreover, at each time step, all robots in the swarm move with different rotational and translational speeds where the translational speed is capped at ($d_{\mt{max}}/\Delta t$).
\newline
\noindent \textbf{Effect of the noisy sensors:} If a measurement is noisy, it implies that the distance a robot estimates using \textbf{\texttt{DistAvoDynObs}} and \textbf{\texttt{DistAttractTarget}} is not accurate, which might cause collisions. To avoid collisions, a simple solution would be to tweak the safety thresholds defined in Section \ref{collAvo}; however, this would require a bounded noise model with known bounds. 
We ran two experiments to analyze the affect of noise on the behavior. 

\begin{figure}[t!]
    \centering
    \includegraphics[width=0.5\linewidth]{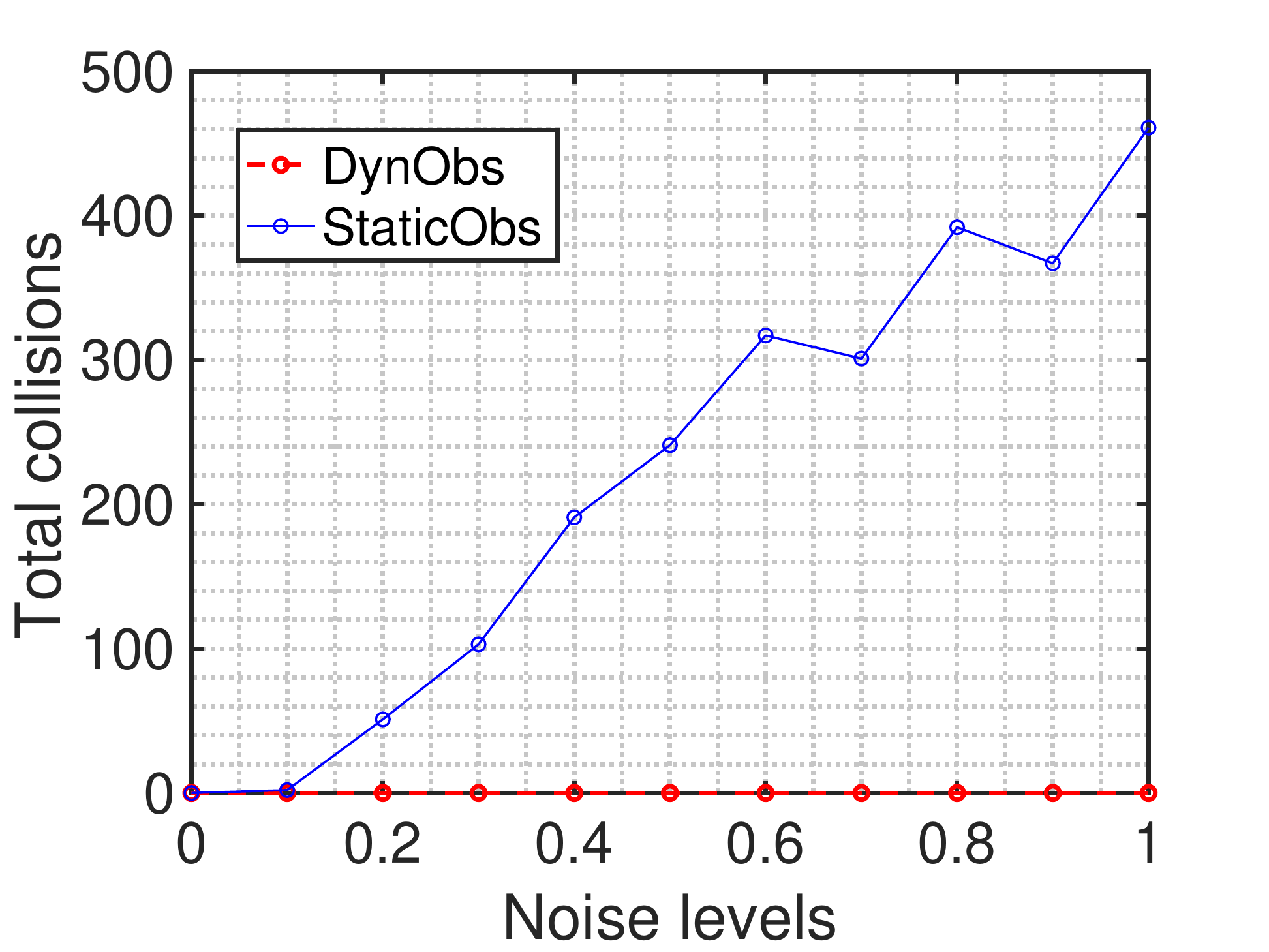}
    \caption{Total collisions with dynamic and static obstacles; safety bounds assume noiseless measurements.
    }
    \label{fig:prob_untrunc}
\end{figure}

\noindent \textbf{Experiment 1:} We did not modify the safety thresholds and all robots implemented algorithm \ref{algo:ca1} as is, with $p=7$. In Fig. \ref{fig:prob_untrunc}, we show how increasing the noise levels (Eq.~\ref{noiseAdd}) affects the total number of collisions observed between robots and static and dynamic obstacles. 
We repeated this experiment (100 simulations per noise level) for each of the following: (i) different white noise added to each sensor on every robot (ii) the same white noise added to all sensors on a robot (iii) the same white noise added to all sensors on all the robots in the swarm.
In all cases of added noise, we observed that the number of collisions with static obstacles (targets and environment boundary) increased with an increase in the noise levels. In contrast, at all noise levels there were no collisions among robots. 

A possible hypothesis for this emergence is that our control algorithm ensures that robots in close vicinity move so as to avoid each other. 
For clusters of moving robots, the virtual source is already closer than the actual source, resulting in less sensitivity to noisy measurements. Such robustness to noise could be attributed to the fact that the randomness of white noise is averaged out to zero in the swarm. In case of static obstacles, since the source does not move away from the robot, the noise filtering is no longer two-sided and hence there is a higher probability of collisions. 

\noindent \textbf{Experiment 2:} Since collisions with only static obstacles were observed, we truncated the noise on static signals to $-0.6 < n_{s}^k \leq 0.6$ $\forall s \in \{g,e\}$, $\forall k \in \{1 \cdots p\}$. We also changed the safety thresholds for static obstacles to $I_s^{\mt{safe}} = I_s^{\mt{safe}}(1-0.6) \forall s \in \{g,e\}$. This ensured that the collision avoidance behavior is triggered in the worst case scenario of a robot being located almost at $r_s^{\mt{safe}}$ of a static source but estimating it to be further away. A decrease in $I_s^{\mt{safe}}$ implies that $r_s^\mt{safe}$ increased, resulting in the increase of $r_g^{\mt{encap}}$ or the area of the annular ring in Fig. \ref{fig:totRobs}. This led to an increase in the bound on $n_0$ from Eq. (\ref{totalRobs}). Given this more conservative bound, we did not observe any collisions with static or dynamic obstacles.

\begin{figure}[t!]
    \centering
    \begin{subfigure}{0.49\linewidth}
    \includegraphics[width=\textwidth]{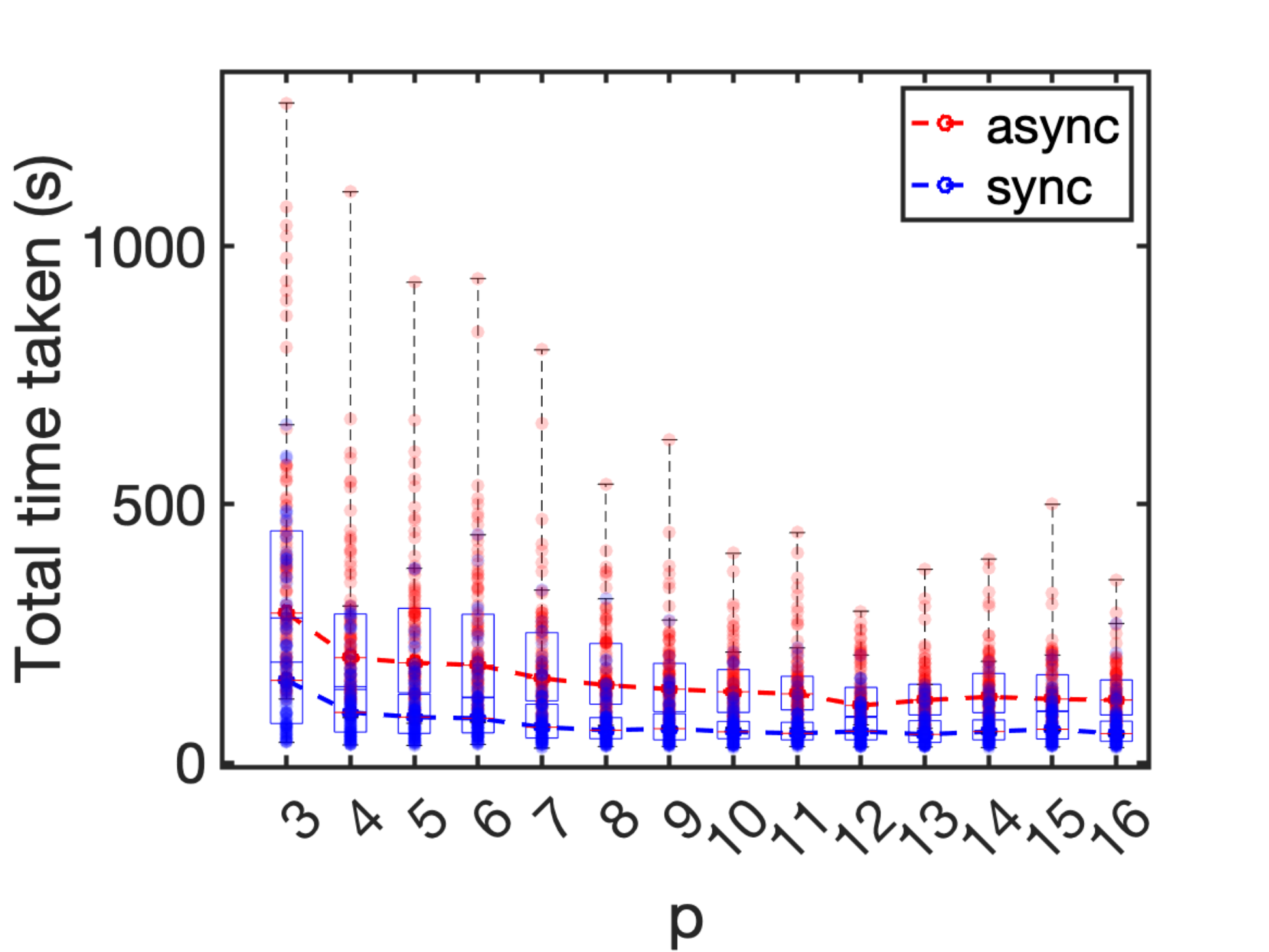}
     \caption{}
      \label{fig:diffFreq_sameStartTime}
    \end{subfigure}
    \begin{subfigure}{0.49\linewidth}
    \includegraphics[width=\textwidth]{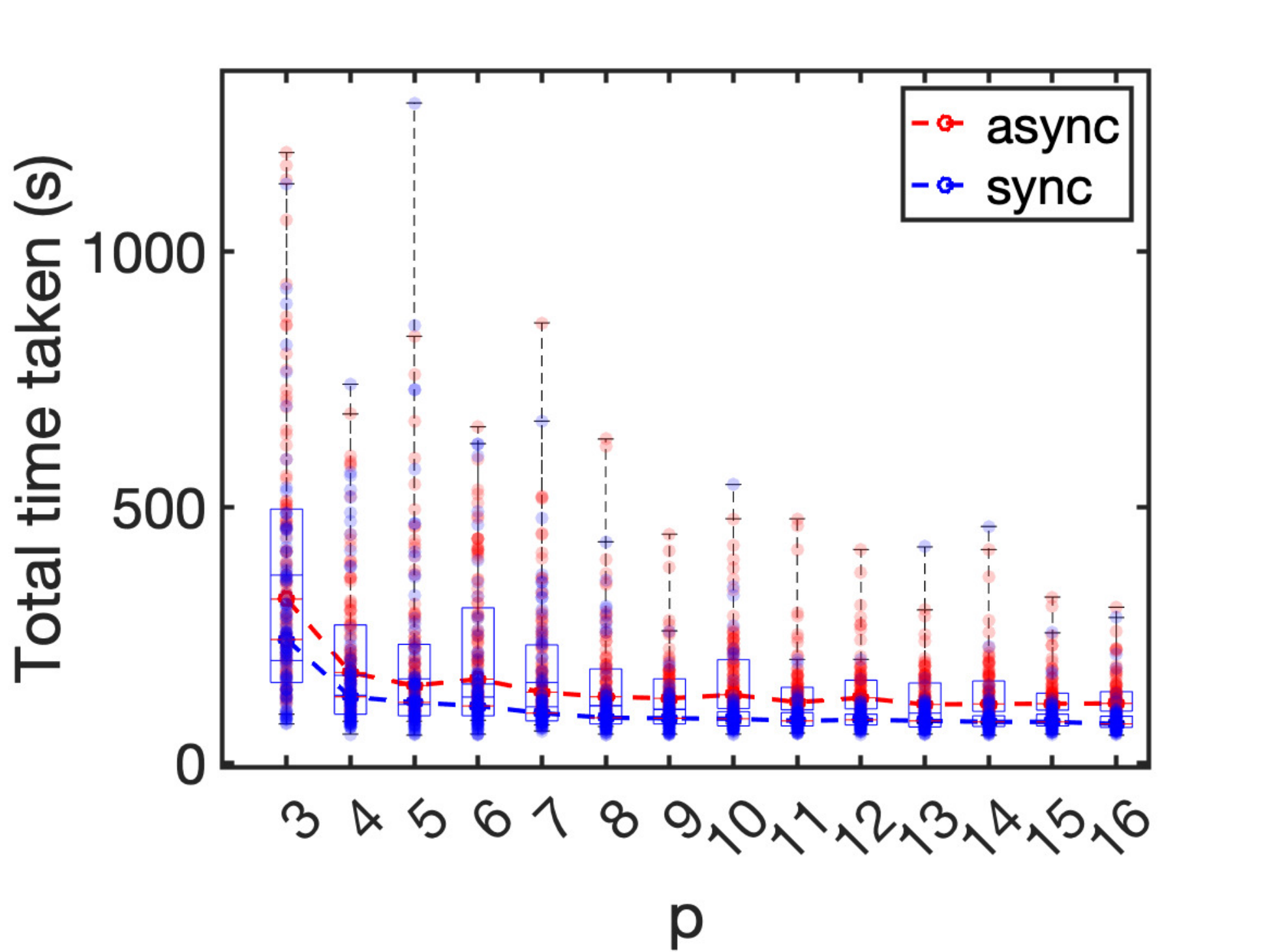}
     \caption{}
      \label{fig:sameFreq_diffStartTime}
    \end{subfigure}
    \vspace{-0.4em}
    \caption{The total time taken for task completion as a function of $p$ for asynchronous and synchronous control such that in (a) each robot has a different sensor update frequency  with the same initial start time and in (b) each robot has a different initial start time with same sensor update frequency. 
    The box plot shows median, 25th and 75th percentiles and the min/max values. The line connects the medians.}
\end{figure}

\begin{figure}[t!]
    \vspace{-1.4em}
    \centering
    \includegraphics[width=0.55\linewidth]{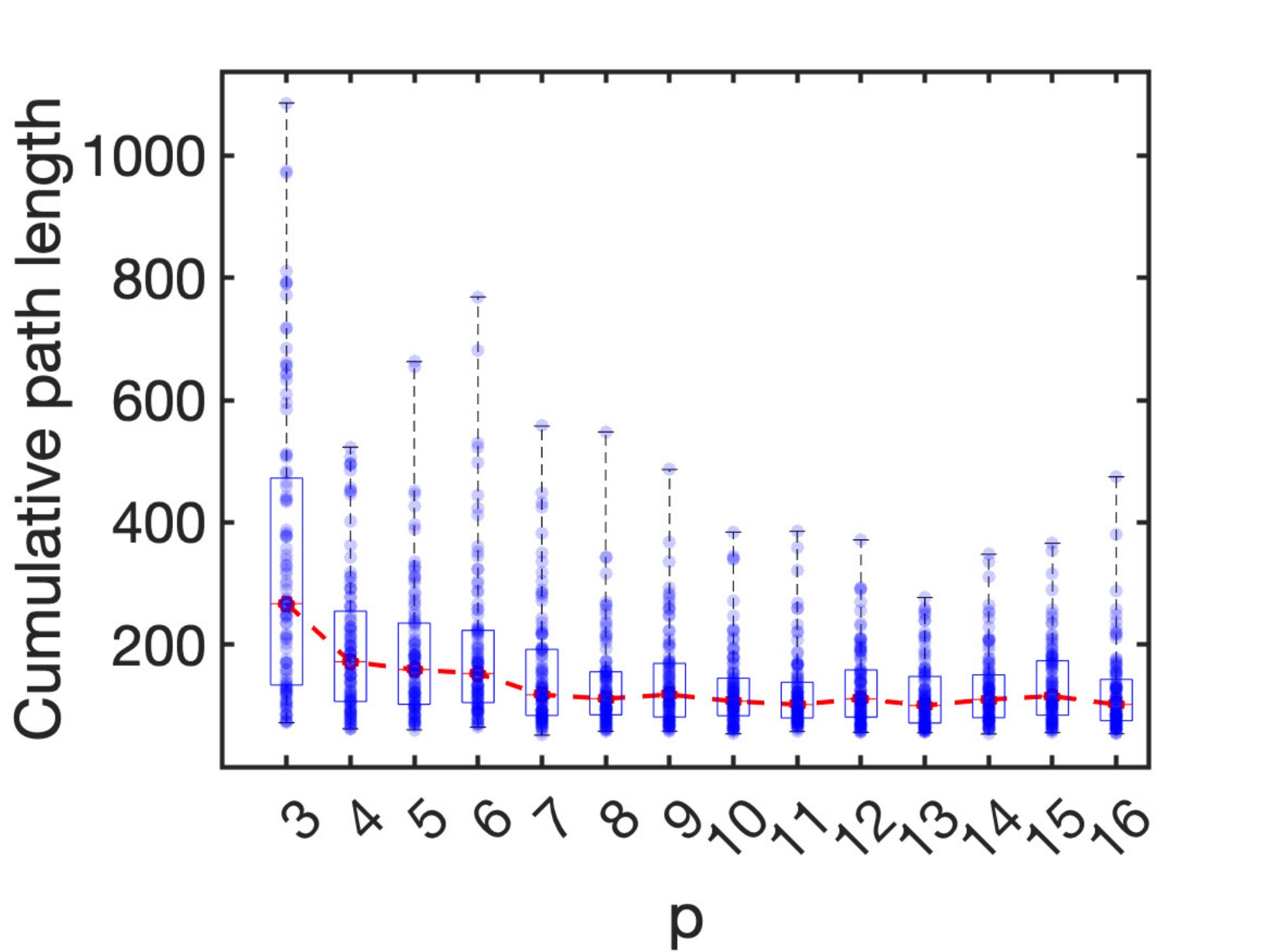}
    \caption{Cumulative path length of all robots as a function of $p$. Distance is measured in robot diameter. 
    The line connects the medians.
    }
    \label{fig:varyNsensors}
\end{figure}

For all the remaining analysis below, we consider a noise level of $15 \%$ in the measurements and a truncated noise addition (within 2 standard deviations) for signals from static sources (targets and the boundary). 
\newline
\noindent \textbf{Effect of asynchronous control:} In Fig. \ref{fig:diffFreq_sameStartTime} we show how the frequency of the sensing and control updates affects the total time taken for encapsulation as compared to synchronous control. We set $n_g = 5$ and an initial update time of $t_0 = 0$ for all robots. For asynchronous control, the sensors update frequency for each robot was randomly chosen from $[1,2,3,4]$ global time steps; it was 1 time step for synchronous control. Less frequent sensor and control updates of some robots in the swarm resulted in a higher total time taken for task completion. We observed a similar phenomenon in  Fig. \ref{fig:sameFreq_diffStartTime} where  the sensor and control update frequency is the same (every 2 global time steps) for all robots, but they differ in their initial start time which was randomly chosen from $\{0,1,2,3,4,5\}$. For synchronous control, all robots started at $t_0 = 0$ with sensing and control updates every 2 global time steps. 
\newline
\noindent \textbf{Effect of the number of sensors $p$:} 
In Fig. \ref{fig:varyNsensors}, we show how increasing the number of sensors on a robot affects the cumulative path length traveled by the robots for encapsulating the target.
Here $n_g = 5$ and the control is synchronous. The cumulative path length decreased quickly as the number of sensors are increased. This is because with an increase in $p$, robots have an increased sense of directionality, resulting in less chattering due to simultaneous attraction and repulsion towards the target and nearby robots.

\begin{figure}[t!]
    \centering
    \begin{subfigure}{0.49\linewidth}
    \includegraphics[width=\textwidth]{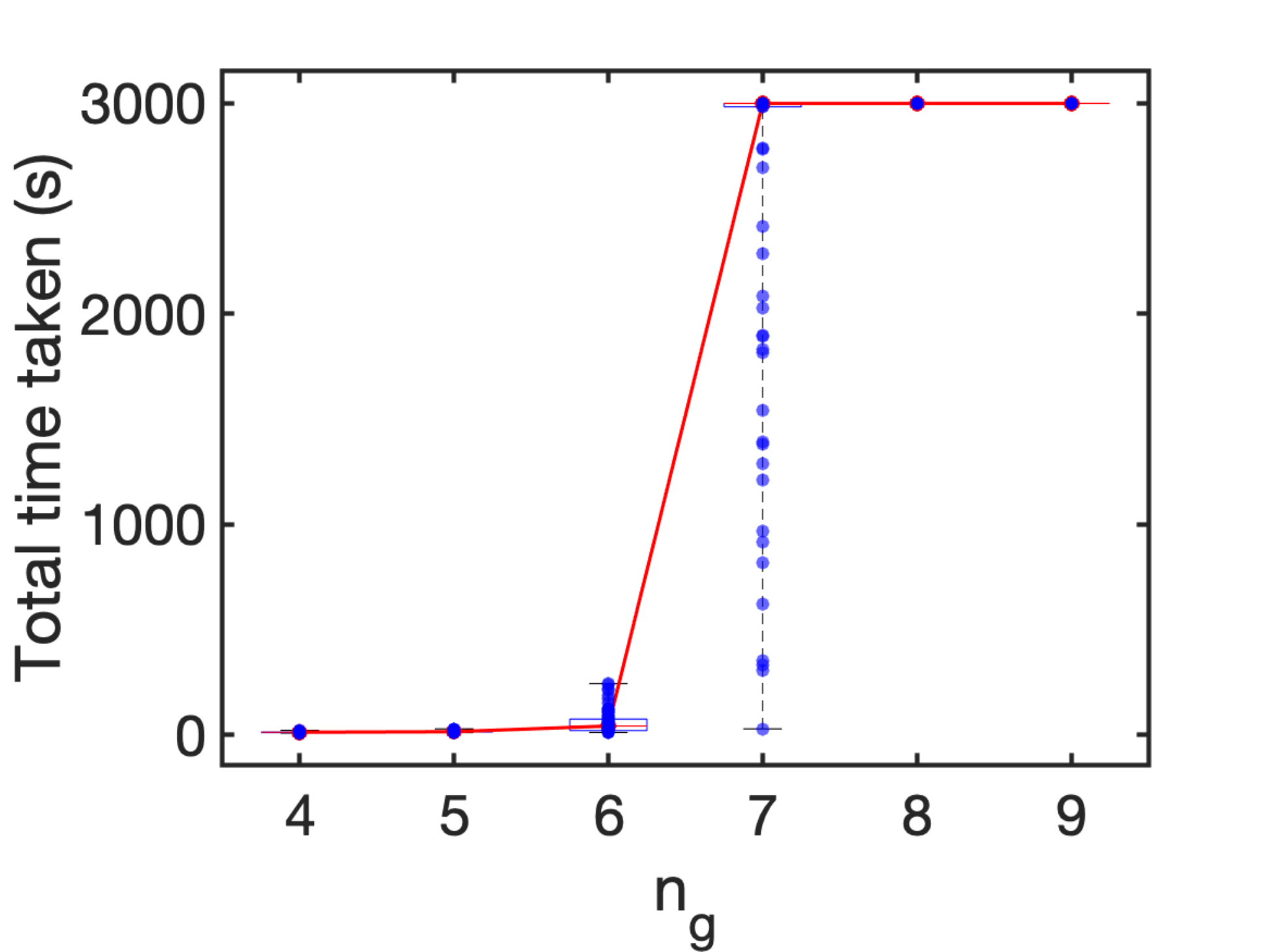}
     \caption{}
      \label{fig:varyNgTime}
    \end{subfigure}
    \begin{subfigure}{0.49\linewidth}
    \includegraphics[width=\textwidth]{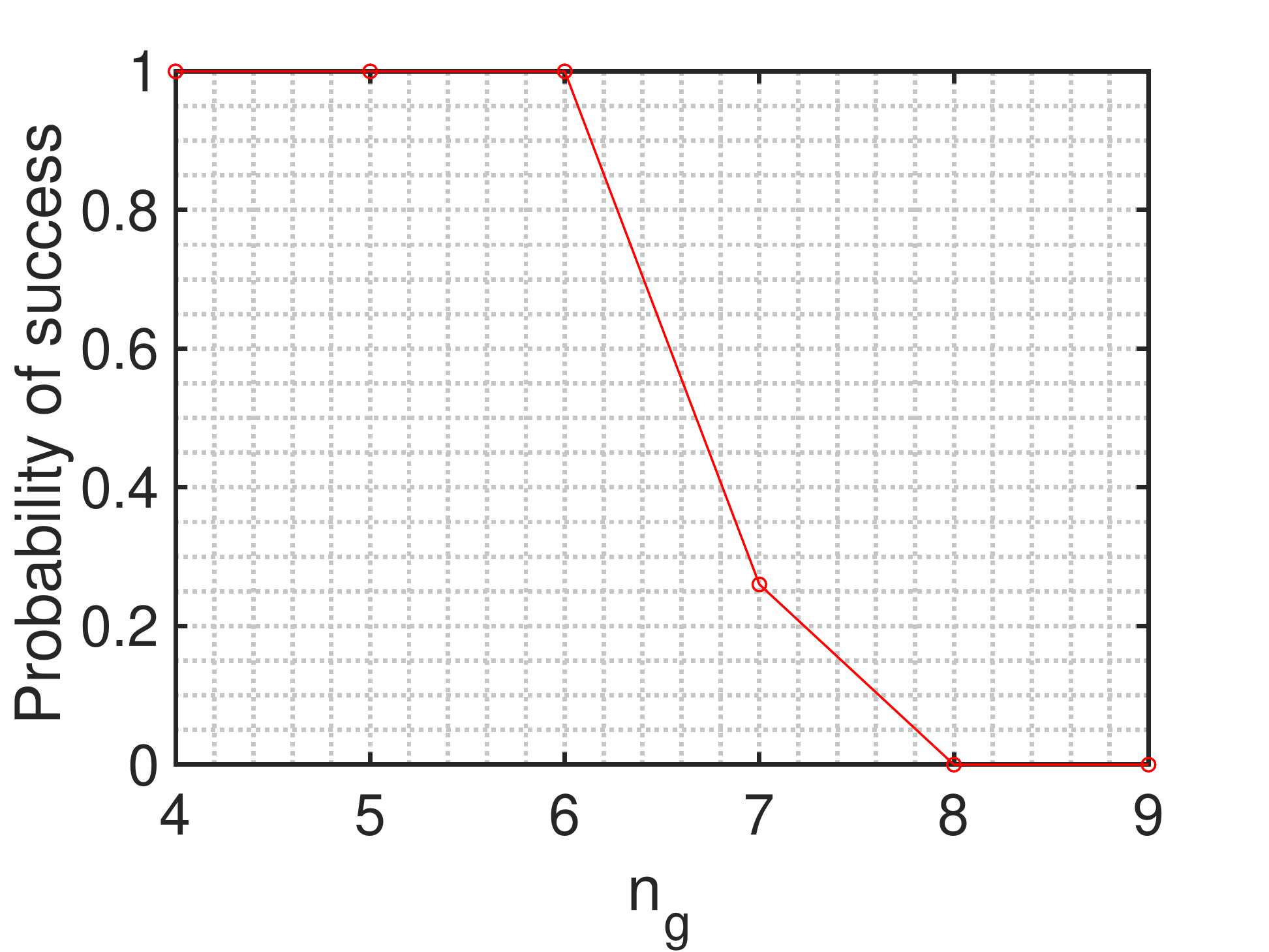}
     \caption{}
      \label{fig:varyNgRateSuc}
    \end{subfigure}
    \vspace{-0.4em}
    \caption{(a) Total time taken for target encapsulation (with simulation time capped at 3000 time-steps) and (b) probability of success for target encapsulation as a function of $n_g$ when $n_0 = 6.9051$.}
\end{figure}

\noindent \textbf{Effect of varying $n_g$:}
In Fig. \ref{fig:varyNgTime}, we show how varying the number of robots required for target encapsulation $n_g$ affects the total time taken for target encapsulation. The number of sensors is $p=8$, and  $n_0 = 6.9051$ implying $n_g \leq 6$ for guaranteed task completion . 
Fig. \ref{fig:varyNgRateSuc} shows the probability of task completion. As expected, the task is completed $100\%$ of the time if $n_g \leq 6$, it is completed $26 \%$ of the time for $n_g = 7$, and is never completed within the time bound for $n_g>7$. Hence, the guaranteed desired emergence only happens when the conditions in Section \ref{proof} are met.

\noindent \textbf{Scalability and Emergent behaviors(video):} In the supplemental video we demonstrate different emergent swarm behaviors created by different parameter choices. 
Our control algorithm is highly scalable, as shown in a large scale simulation of 25 targets and 200 robots. We also show task completion for asymmetric placement of sensors. 

\section{Conclusion}
In this paper, we show how robots equipped with simple omnidirectional sensors and an isotropic signal emitter are capable of finding and encapsulating targets in the environment while avoiding both static and dynamic obstacles. Our decentralized controller is agnostic to the number of robots and targets in the environment. We presented a detailed analysis of the implemented decentralized controller and provided bounds on the maximum step size of a robot, minimum number of required sensors, and maximum number of robots required for encapsulation such that the desired emergent behavior is guaranteed. We further studied the effects of noise, uncertainties, and asynchronous control on the emergent behavior and observed an interesting phenomenon of robot-robot collision avoidance regardless of the (bounded) noise level.
In future work, we will implement our algorithm on physical robots. We will analyze how the presence of other obstacles and targets moving in patterns affect the emergent behavior.
We will also explore different target search strategies such as Le{\'v}y walk and correlated random walks for a swarm with no memory.

\bibliography{references}
\bibliographystyle{ieeetr}

\end{document}